\def\year{2018}\relax
\documentclass[letterpaper]{article} 
\usepackage{aaai18,mathpazo}  
\usepackage{times}  
\usepackage{helvet}  
\usepackage{courier}  
\usepackage{url}  
\usepackage{graphicx,amsmath}  
\begin{document}
%
\title{Probabilistic Planning by Probabilistic Programming}
\author{Vaishak Belle\\
University of Edinburgh \& Alan Turing Institute\\
{  vaishak@ed.ac.uk}}
\maketitle
\begin{abstract}
Automated planning is a major topic of research in artificial intelligence, and enjoys a long and distinguished history. The classical paradigm assumes a distinguished initial state, comprised of a set of facts, and is defined over a set of actions which change that state in one way or another. Planning in many real-world settings, however, is much more involved: an agent's knowledge is almost never simply a set of facts that are true, and actions that the agent intends to execute never operate the way they are supposed to. Thus, probabilistic planning attempts to incorporate stochastic models directly into the planning process. In this article, we briefly report on probabilistic planning through the lens of probabilistic programming: a programming paradigm that aims to ease the specification of structured probability distributions. In particular, we provide an overview of the features of two systems, HYPE and ALLEGRO, which emphasise different strengths of probabilistic programming that are particularly useful for complex modelling issues raised in probabilistic planning. Among other things, with these systems, one can instantiate planning problems with growing and shrinking state spaces, discrete and continuous probability distributions, and non-unique prior distributions in a first-order setting. 
\end{abstract}

\section{Introduction}

Automated planning is a major topic of research in artificial intelligence, and enjoys a long and distinguished history \cite{strips:-a-new-approach-to-the-application-of-theorem}. The classical paradigm assumes a distinguished initial state, comprised of a set of facts, and is defined over a set of actions which change that state in one way or another. Actions are further characterised in terms of their applicability conditions, that is, things that must be true for the agent to be able to execute it, and effects, which procedurally amounts to adding new facts to a state while removing others. The scientific agenda is then to design algorithms that synthesise a sequence of actions that takes the agent from an initial state to a desired goal state. 

From the early days, automated planning was motivated by robotics applications. But it was observed that the real world -- or more precisely, the robot's knowledge about the world -- is almost never simply a set of facts that are true, and actions that the agent intends to execute never operate the way they are supposed to. One way to make sense of this complication is to separate the ``high-level reasoning,'' in our case the planner's search space, from the low-level sensor-motor details. On the positive side, such a move allows the plan representation to be finite, discrete and simple. On the negative side, significant expert knowledge has to go into materialising this separation of concerns, possibly at the loss of clarity on the behaviour of the system as a whole.

Incidentally, by testing the robot's effectors repeatedly in a controlled environment, one can approximate the uncertain effects of an action in terms of a probability distribution. Similarly, based on minimalistic assumptions about the environment, expressed as a probabilistic prior, by repeated sampling, the robot can update its prior to converge on a reasonable posterior that approximates the environment \cite{probabilistic-robotics}. To that end, probabilistic planning attempts to incorporate such models directly into the planning process. There are to-date numerous languages and algorithmic frameworks for probabilistic planning, e.g., \cite{probabilistic-planning-via-heuristic-forward,decision-theoretic-planning:-structural-assumptions,planning-and-acting-in-partially-observable,planning-under-uncertainty-for-robotic}. 

In this article, we briefly report on probabilistic planning through the lens of {\it probabilistic programming}  \cite{probabilistic-programming}. Probabilistic programming is a programming paradigm that aims to ease the specification of structured probability distributions;  these languages are developed so as to enable modularity and re-use in probabilistic machine learning applications. Their atomic building blocks incorporate stochastic primitives, and the  formal representation also allows for compositionality. Here, we specifically provide an overview of the features of two kinds of systems, both of which have their roots in logic programming: 

\begin{itemize}

\item HYPE \cite{planning-in-discrete-and-continuous-markov}: a planning framework based on {\it distributional clauses} \cite{the-magic-of-logical-inference-in-probabilistic}; and 

\item {ALLEGRO} \cite{allegro:-belief-based-programming-in-stochastic}: a high-level control programming framework that extends GOLOG \cite{knowledge-in-action:-logical-foundations}.

\end{itemize}

These two systems emphasise different strengths of probabilistic programming, which we think are particularly useful for complex modelling issues raised in probabilistic planning. HYPE can easily describe growing and shrinking state spaces owing to uncertainty about the existence of objects, and thus is closely related to BLOG models \cite{blog:-probabilistic-models-with:book,first-order-open-universe-pomdps}. Since HYPE is an extension of PROBLOG \cite{problog:-a-probabilistic-prolog-and-its-application}, it stands to benefit from a wide range of applications and machine learning models explored with PROBLOG.\footnote{\small  dtai.cs.kuleuven.be/problog} The dynamical aspects of the domain are instantiated by reifying time as an argument in the predicates, and so it is perhaps most appropriate for finite horizon planning problems. 

ALLEGRO treats actions as first-class citizens and is built on a rich model of dynamics and subjective probabilities, which allows it to handle context-sensitive effect axioms, and non-unique probability measures placed on first-order formulas. GOLOG has also been widely used for a range of applications that apply structured knowledge (e.g., ontologies) in dynamical settings \cite{cognitive-robotics}, and ALLEGRO stands to inherit these developments. GOLOG has also been shown as a way to structure search in large plan spaces \cite{exploiting-procedural-domain-control}. Finally, since there are constructs for iteration and loops, such programs are most appropriate for modelling non-terminating behaviour \cite{a-logic-for-non-terminating-golog-programs}.

In the sequel, we describe the essential formal and algorithmic contributions of these systems before concluding with open computational issues. 

\section{HYPE}

PROBLOG aims to unify logic programming and probabilistic specifications, in the sense of providing a language to specify distributions together with the means to query about the probabilities of events. As a very simple example, to express that the object $  c$ is  on the table with a certain probability, and that all objects on the table are also in the room, we would write (free variables are assumed to be quantified from the outside): 

\begin{align*}  
  .6::onTable(c). \\
  inRoom(x) \leftarrow onTable(x). 
\end{align*}

This then allows us to query the probability of  atoms such $  inRoom(c)$. 

A more recent extension \cite{the-magic-of-logical-inference-in-probabilistic} geared for continuous distributions and other infinite event-space distributions allows the head atom of a logical rule to be drawn from a distribution directly, by means of the following syntax: 

\[  h \sim D \leftarrow b_1, \ldots, b_n. \]

For example, suppose there is an urn with an unknown number of balls \cite{blog:-probabilistic-models-with:book}. Suppose we pull a ball at a time and put it back in the urn, and repeat these steps (say) 6 times. Suppose further we have no means of identifying if the balls drawn were distinct from each other. A probabilistic program for this situation might be as follows: 

\begin{align*}
  n \sim poisson(6). \\
  pos(x) \sim uniform(1,10) \leftarrow between(1, \simeq\!(n),x).  
\end{align*}

For simplicity, we assume here that the physical form of the urn is a straight line of length 10, and the position of a ball is assumed to be anywhere along this line. 

HYPE is based on a dynamic extension that allows us to temporally index the truth of atoms, and so can be used to reason about actions. For example, the program: 

\begin{align*}
  numBehind(x,t+1) \sim poisson(1) \leftarrow removeObj(x,t). 
\end{align*}

says that on removing the object $  x$ at $  t$, we may assume that there are objects -- typically one such object -- behind $  x$. Such programs can be used in object tracking applications to reason about occluded objects \cite{hybrid-probabilistic-logic-programming}. 

A common declaration in many robotics applications \cite{probabilistic-robotics} is to define actions and sensors with an error profile, such as a Gaussian noise model. These can be instantiated in HYPE using: 

\begin{align*}
  pos(x, t+1) \sim gaussian(\simeq\!(pos(x,t)) + 1, var) \\ \quad \qquad    \leftarrow move(x,t). \\ 
  obs(x,t+1) \sim gaussian(\simeq \! (pos(x,t)), var). 
\end{align*}

The first rule says that the position of $  x$ on doing a move action is drawn from a normal distribution whose mean is $  x$'s current position incremented by one. The second one says that observing the current position of $  x$ is subject to additive Gaussian noise. 

As an automated planning system, HYPE instantiates a Markov decision process (MDP) \cite{markov-decision-processes:-discrete}. Recall that MDPs are defined in terms of states, actions, stochastic transitions and reward functions, which can be realised in the above syntax using rules such as: 

\begin{align*}
  poss(act, t) \leftarrow conditions(t). \\
  reward(num, t) \leftarrow conditions(t). 
\end{align*}

To compute a policy, which is a mapping from states and time points to actions, HYPE combines importance sampling and SLD resolution to effectively bridge the high-level symbolic specification and the probabilistic components of the programming model. HYPE allows states and actions to be discrete or continuous, yielding a very general planning system. Empirical evaluations are reported in \cite{planning-in-discrete-and-continuous-markov} and \cite{hybrid-relational-mdps}. 

\section{ALLEGRO}

The GOLOG language has been successfully used in a wide range of applications involving control and planning \cite{cognitive-robotics}, 
and is based on a simple ontology that all changes are a result of named actions \cite{knowledge-in-action:-logical-foundations}. 
An initial state describes the truth values of properties, and actions may affect these values in non-trivial context-sensitive ways. 
In particular, GOLOG is a programming model where executing actions are the simplest instructions in the program, upon which more involved constructions for iteration and loops are defined. 
For example, a program to clear a table containing an unknown number of blocks would be as follows: 

\[([\pi x ~onTable(x)?; removeObj(x)] )^*; \neg \exists x ~ onTable(x)? \]
Here, $ \pi$ is the non-deterministic choice of argument, semi-colon denotes sequence, ? allows for test conditions, and $*$ is unbounded iteration.  The program terminates successfully  because the sub-program before the final test condition removes every object from the table. 

As argued in \cite{cognitive-robotics}, the rich syntax of GOLOG allows us, on the one hand, to represent policies and plans in an obvious fashion; for example: 

\[  a_1; \ldots; a_n; P?\]
ensures that the goal $P$ is true on executing the sequence of actions. However, the syntax also allows open-ended search; for example: 

\[while~~ \neg P~~\pi a.~a \]
tries actions until $P$ is made true. 
The benefit of GOLOG then is that it allows us to explore plan formulations between these two extremes, including partially specified programs that are completed by a meta-language planner. 

ALLEGRO augments the underlying ontology to reason about probability distributions over state properties, and allow actions with uncertain (stochastic) effects. In logical terms, the semantical foundations rests on a rich logic of belief and actions. Consequently, it can handle partial probabilistic specifications. For example, one can say $  c$ is on the table with a certain probability as before: \(  pr(onTable(c)) = .6\), but it is also possible to express the probability that there is an object on the table without knowing which one: $  pr(\exists x~onTable(x)) = .6$. We can go further and simply say that there is a non-zero probability of that statement: $  pr(\exists x~onTable(x)) > 0$, which means that any distribution satisfying the formula is admissible. Such a feature can be very useful: for example, in \cite{integrated-task-and-motion-planning}, it is argued that when planning in highly stochastic  environments, it is useful to allow a margin of error in the probability distributions defined over state properties. 

To model the case of Gaussian error models, actions with uncertain effects are given a general treatment. For one thing, the effects of actions are axiomatised using the notion of successor state axioms which incorporate Reiter's solution to the frame problem \cite{knowledge-in-action:-logical-foundations}. So, for example, changing the position of an object using a move action can be expressed as: 

\begin{align*}  pos(x, do(a,s)) = u \equiv \\ \qquad    (a = move(x,y) \land pos(x,s) = u+y) \\ 
\qquad   \lor (a\neq move(x,y) \land pos(x,s) = u)\end{align*}

This says that if the action of moving $  x$ was executed, its position (along a straight line) is decremented by $  y$ units, and for all other actions, the position is unaffected. To deal with uncertain effects, we will distinguish between what the agent intends and what actually happens. That is, let $  move(x,y,z)$ be a new action type, where $  y$ is what the agent intends, and $  z$ is what happens. Then, the successor state axiom is rewritten as follows: 

\begin{align*}  pos(x, do(a,s)) = u \equiv \\ \qquad   (a = move(x,y,z) \land pos(x,s) = u+z) \\ 
\qquad   \lor (a\neq move(x,y,z) \land pos(x,s) = u)\end{align*}
The story remains essentially the same, except that $  z$ determines the actual position in the successor state, but it is not in control of the agent. A Gaussian error profile can be accorded to this action by means: 

\[  l(move(x,y,z),s) = gaussian(z; y, var)\]

That is, the actual value is drawn from a Gaussian whose mean is the intended argument $  y$. Analogously, attributing additive Gaussian noise in a sensor for observing the position is defined using: 

\[  l(obs(x,z),s) = gaussian(z; pos(x,s), var) \]
That is, the observation $  z$  is drawn from a Gaussian whose mean is the actual position of the object $  x$. 

As hinted above, as an extension to GOLOG, the syntax of ALLEGRO is designed to compactly represent full or partial plans and policies in a general way, and on termination, ALLEGRO programs can be tested for any probabilistic or expectation-based criteria. 
The foundations of ALLEGRO was established in \cite{allegro:-belief-based-programming-in-stochastic}
with a discussion on its empirical behaviour against a predecessor based on goal regression. 

\section{Conclusions}

Automated planning is often deployed in an application context, and in highly stochastic and uncertain domains, the planning model may be derived from a complex learning and reasoning pipeline, or otherwise defined over non-trivial state spaces with unknowns. In this article, we reported on two probabilistic programming systems to realise such pipelines. Indeed, combining automated planning and probabilistic programming is receiving considerable attention recently, e.g., \cite{first-order-open-universe-pomdps}. 
These languages are general purpose, and their first-order expressiveness can not only enable a compact codification of the domain but also achieve computational leverage. 

One of the key concerns with the use of probabilistic programming and stochastic specifications generally is that most systems perform inference by Monte Carlo sampling. 
As is well-known, one is able to only obtain asymptotic guarantees with such methods, and moreover,  handling low-probability observations can be challenging. In that regard, there have been recent logical approaches for inferring in mixed discrete-continuous probability spaces with tight bounds on the computed answers \cite{hashing-based-approximate-probabilistic-inference,probabilistic-inference-in-hybrid-domains,approximate-counting-in-smt-and-value-estimation}. Since HYPE, ALLEGRO and many such systems use probabilistic inference as a fundamental computational backbone, the question then is whether the aforementioned  approaches can enable robust planning and programming frameworks in stochastic domains.

\bibliographystyle{aaai}
\bibliography{group}

\begin{thebibliography}{}

\bibitem[\protect\citeauthoryear{Baier, Fritz, and
  McIlraith}{2007}]{exploiting-procedural-domain-control}
Baier, J.~A.; Fritz, C.; and McIlraith, S.~A.
\newblock 2007.
\newblock Exploiting procedural domain control knowledge in state-of-the-art
  planners.
\newblock In {\em Proc. ICAPS},  26--33.

\bibitem[\protect\citeauthoryear{Belle and
  Levesque}{2015}]{allegro:-belief-based-programming-in-stochastic}
Belle, V., and Levesque, H.~J.
\newblock 2015.
\newblock Allegro: Belief-based programming in stochastic dynamical domains.
\newblock In {\em IJCAI}.

\bibitem[\protect\citeauthoryear{Belle, Passerini, and Van~den
  Broeck}{2015}]{probabilistic-inference-in-hybrid-domains}
Belle, V.; Passerini, A.; and Van~den Broeck, G.
\newblock 2015.
\newblock Probabilistic inference in hybrid domains by weighted model
  integration.
\newblock In {\em IJCAI}.

\bibitem[\protect\citeauthoryear{Belle, Van~den Broeck, and
  Passerini}{2015}]{hashing-based-approximate-probabilistic-inference}
Belle, V.; Van~den Broeck, G.; and Passerini, A.
\newblock 2015.
\newblock Hashing-based approximate probabilistic inference in hybrid domains.
\newblock In {\em UAI}.

\bibitem[\protect\citeauthoryear{Boutilier, Dean, and
  Hanks}{1999}]{decision-theoretic-planning:-structural-assumptions}
Boutilier, C.; Dean, T.; and Hanks, S.
\newblock 1999.
\newblock Decision-theoretic planning: Structural assumptions and computational
  leverage.
\newblock {\em Journal of Artificial Intelligence Research} 11(1):94.

\bibitem[\protect\citeauthoryear{Chistikov, Dimitrova, and
  Majumdar}{2015}]{approximate-counting-in-smt-and-value-estimation}
Chistikov, D.; Dimitrova, R.; and Majumdar, R.
\newblock 2015.
\newblock Approximate counting in smt and value estimation for probabilistic
  programs.
\newblock In {\em TACAS}, volume 9035.
\newblock  320--334.

\bibitem[\protect\citeauthoryear{Cla{\ss}en and
  Lakemeyer}{2008}]{a-logic-for-non-terminating-golog-programs}
Cla{\ss}en, J., and Lakemeyer, G.
\newblock 2008.
\newblock A logic for non-terminating golog programs.
\newblock In {\em KR},  589--599.

\bibitem[\protect\citeauthoryear{Domshlak and
  Hoffmann}{2007}]{probabilistic-planning-via-heuristic-forward}
Domshlak, C., and Hoffmann, J.
\newblock 2007.
\newblock Probabilistic planning via heuristic forward search and weighted
  model counting.
\newblock {\em JAIR} 30:565--620.

\bibitem[\protect\citeauthoryear{Fikes and
  Nilsson}{1971}]{strips:-a-new-approach-to-the-application-of-theorem}
Fikes, R., and Nilsson, N.~J.
\newblock 1971.
\newblock {STRIPS}: A new approach to the application of theorem proving to
  problem solving.
\newblock In {\em Proc. IJCAI},  608--620.

\bibitem[\protect\citeauthoryear{Gordon \bgroup et al\mbox.\egroup
  }{2014}]{probabilistic-programming}
Gordon, A.~D.; Henzinger, T.~A.; Nori, A.~V.; and Rajamani, S.~K.
\newblock 2014.
\newblock Probabilistic programming.
\newblock In {\em Proc. International Conference on Software Engineering}.

\bibitem[\protect\citeauthoryear{Gutmann \bgroup et al\mbox.\egroup
  }{2011}]{the-magic-of-logical-inference-in-probabilistic}
Gutmann, B.; Thon, I.; Kimmig, A.; Bruynooghe, M.; and De~Raedt, L.
\newblock 2011.
\newblock The magic of logical inference in probabilistic programming.
\newblock {\em TPLP} 11:663--680.

\bibitem[\protect\citeauthoryear{Kaelbling and
  Lozano-P{\'e}rez}{2013}]{integrated-task-and-motion-planning}
Kaelbling, L.~P., and Lozano-P{\'e}rez, T.
\newblock 2013.
\newblock Integrated task and motion planning in belief space.
\newblock {\em I. J. Robotic Res.} 32(9-10):1194--1227.

\bibitem[\protect\citeauthoryear{Kaelbling, Littman, and
  Cassandra}{1998}]{planning-and-acting-in-partially-observable}
Kaelbling, L.~P.; Littman, M.~L.; and Cassandra, A.~R.
\newblock 1998.
\newblock Planning and acting in partially observable stochastic domains.
\newblock {\em Artificial Intelligence} 101(1--2):99 -- 134.

\bibitem[\protect\citeauthoryear{Lakemeyer and
  Levesque}{2007}]{cognitive-robotics}
Lakemeyer, G., and Levesque, H.~J.
\newblock 2007.
\newblock {Cognitive robotics}.
\newblock In {\em Handbook of Knowledge Representation}. Elsevier.
\newblock  869--886.

\bibitem[\protect\citeauthoryear{Milch \bgroup et al\mbox.\egroup
  }{2007}]{blog:-probabilistic-models-with:book}
Milch, B.; Marthi, B.; Russell, S.; Sontag, D.; Ong, D.; and Kolobov, A.
\newblock 2007.
\newblock {BLOG}: Probabilistic models with unknown objects.
\newblock {\em Introduction to statistical relational learning}  373.

\bibitem[\protect\citeauthoryear{Nitti, Belle, and
  Raedt}{2015}]{planning-in-discrete-and-continuous-markov}
Nitti, D.; Belle, V.; and Raedt, L.~D.
\newblock 2015.
\newblock Planning in discrete and continuous markov decision processes by
  probabilistic programming.
\newblock In {\em ECML}.

\bibitem[\protect\citeauthoryear{Nitti \bgroup et al\mbox.\egroup
  }{2017}]{hybrid-relational-mdps}
Nitti, D.; Belle, V.; De~Laet, T.; and De~Raedt, L.
\newblock 2017.
\newblock Planning in hybrid relational mdps.
\newblock {\em Machine Learning}  1--28.

\bibitem[\protect\citeauthoryear{Nitti}{2016}]{hybrid-probabilistic-logic-programming}
Nitti, D.
\newblock 2016.
\newblock {\em Hybrid Probabilistic Logic Programming}.
\newblock Ph.D. Dissertation, KU Leuven.

\bibitem[\protect\citeauthoryear{Ong \bgroup et al\mbox.\egroup
  }{2010}]{planning-under-uncertainty-for-robotic}
Ong, S. C.~W.; Png, S.~W.; Hsu, D.; and Lee, W.~S.
\newblock 2010.
\newblock Planning under uncertainty for robotic tasks with mixed
  observability.
\newblock {\em Int. J. Rob. Res.} 29(8):1053--1068.

\bibitem[\protect\citeauthoryear{Puterman}{1994}]{markov-decision-processes:-discrete}
Puterman, M.~L.
\newblock 1994.
\newblock {\em Markov Decision Processes: Discrete Stochastic Dynamic
  Programming}.
\newblock New York, NY, USA: John Wiley \& Sons, Inc., 1st edition.

\bibitem[\protect\citeauthoryear{Raedt, Kimmig, and
  Toivonen}{2007}]{problog:-a-probabilistic-prolog-and-its-application}
Raedt, L.~D.; Kimmig, A.; and Toivonen, H.
\newblock 2007.
\newblock Problog: {A} probabilistic prolog and its application in link
  discovery.
\newblock In {\em Proc. IJCAI},  2462--2467.

\bibitem[\protect\citeauthoryear{Reiter}{2001}]{knowledge-in-action:-logical-foundations}
Reiter, R.
\newblock 2001.
\newblock {\em {Knowledge in action: logical foundations for specifying and
  implementing dynamical systems}}.
\newblock {MIT} Press.

\bibitem[\protect\citeauthoryear{Srivastava \bgroup et al\mbox.\egroup
  }{2014}]{first-order-open-universe-pomdps}
Srivastava, S.; Russell, S.~J.; Ruan, P.; and Cheng, X.
\newblock 2014.
\newblock First-order open-universe pomdps.
\newblock In {\em UAI},  742--751.

\bibitem[\protect\citeauthoryear{Thrun, Burgard, and
  Fox}{2005}]{probabilistic-robotics}
Thrun, S.; Burgard, W.; and Fox, D.
\newblock 2005.
\newblock {\em Probabilistic Robotics}.
\newblock {MIT Press}.

\end{thebibliography}

\end{document}